\documentclass[conference]{IEEEtran}
\IEEEoverridecommandlockouts
\usepackage{cite}
\usepackage{amsmath,amssymb,amsfonts}
\usepackage{algorithmic}
\usepackage{graphicx}
\usepackage{textcomp}
\usepackage{xcolor}
\usepackage{booktabs}   
\usepackage{multirow}   
\usepackage{soul}
\usepackage{makecell}

\def\BibTeX{{\rm B\kern-.05em{\sc i\kern-.025em b}\kern-.08em
    T\kern-.1667em\lower.7ex\hbox{E}\kern-.125emX}}
\begin{document}


\title{MMSense: Adapting Vision-based Foundation Model for Multi-task Multi-modal Wireless Sensing\\

}

\author{
    \IEEEauthorblockN{Zhizhen Li\IEEEauthorrefmark{1}, Xuanhao Luo\IEEEauthorrefmark{1}, Xueren Ge\IEEEauthorrefmark{2}, Longyu Zhou\IEEEauthorrefmark{3}, Xingqin Lin\IEEEauthorrefmark{4}, Yuchen Liu\IEEEauthorrefmark{1}}
    
    \IEEEauthorblockA{\IEEEauthorrefmark{1}North Carolina State University, USA; \IEEEauthorrefmark{2}University of Virginia, USA;}
    \IEEEauthorblockA{\IEEEauthorrefmark{3}Singapore University of Technology and Design; \IEEEauthorrefmark{4}NVIDIA Corporation, USA}
    
}

\maketitle

\begin{abstract}
Large AI models have been widely adopted in wireless communications for channel modeling, beamforming, and resource optimization. However, most existing efforts remain limited to single-modality inputs and channel-specific objectives, overlooking the broader potential of large foundation models for unified wireless sensing. To bridge this gap, we propose MMSense, a multi-modal, multi-task foundation model that jointly addresses channel-centric, environment-aware, and human-centered sensing. Our framework integrates image, radar, LiDAR, and textual data by transforming them into vision-compatible representations, enabling effective cross-modal alignment within a unified feature space. A modality gating mechanism adaptively fuses these representations, while a vision-based large language model backbone enables unified feature alignment and instruction-driven task adaptation. Furthermore, task-specific sequential attention and uncertainty-based loss weighting mechanisms enhance cross-task generalization. Experiments on real wireless scenario datasets show that our approach outperforms both task-specific and large-model baselines, confirming its strong generalization across heterogeneous sensing tasks.
\end{abstract}

\section{Introduction}
With the evolution toward 6G wireless networks, artificial intelligence (AI) has emerged as a foundational enabler for achieving ubiquitous connectivity, integrated sensing and communication (ISAC), and intelligent network management. Future network infrastructures are expected not only to provide ultra-reliable and low-latency communication but also to enable advanced capabilities such as environmental perception, human-centric sensing, and autonomous mobile computing. These new demands substantially increase the complexity of communication system design, calling for unified and predictive frameworks that can seamlessly generalize across dynamic environments and heterogeneous tasks.

To enable the perception of environmental dynamics, user activities, and spatial context beyond conventional data transmission, many studies have explored AI-driven approaches for wireless sensing and perception. Leveraging the power of deep learning, researchers have developed diverse data-driven frameworks to enhance channel understanding and signaling awareness. Unlike traditional model-driven methods that rely on simplified physical assumptions, AI-based techniques can capture complex nonlinear relationships and adapt to highly dynamic propagation environments. For instance, \cite{yang2020deep} proposed a deep transfer learning framework employing fully connected neural networks to transfer knowledge from uplink to downlink channels, enhanced with meta-learning for rapid adaptation under limited data. In \cite{li2023spatial}, a spatial–temporal architecture combining graph neural networks (GNNs) and long short-term memory (LSTM) models was introduced to predict link quality under dynamic blockage conditions. Similarly, \cite{li2025bfmloc} transformed channel state information (CSI) into the angular domain and applied a vision Transformer for localization, while \cite{zhang2022deep} designed a customized deep neural network (DNN) for low-complexity precoding in MU-MIMO-OFDM systems, achieving near-weighted minimum mean square error performance with substantially lower computational cost. These advances demonstrate that deep learning offers significant advantages for wireless sensing and communication, providing both efficiency and scalability compared with traditional model-based approaches. However, current methods are highly task-specific—developed  for channel estimation, prediction, localization, or precoding, which restricts cross-task applicability and slows progress toward more unified solutions.

Recently, the paradigm of foundation models (FMs) and large language models (LLMs) has demonstrated remarkable impact in the areas of natural language processing and computer vision. By pretraining on large-scale datasets and fine-tuning with minimal task-specific data, FMs exhibit strong multi-task and few-shot learning capabilities. Inspired by this, several pioneering works have adapted LLMs for wireless domains. For instance, \cite{liu2024llm4cp} proposed an LLM-based channel prediction framework that demonstrates substantially enhanced few-shot generalization capability over different scenarios. 
However, these related studies remain confined to solving individual sensing or communication tasks, without harnessing the task-level generalization and cross-domain adaptability that FMs can offer. Motivated by this gap, a few recent efforts have begun exploring multi-task large models capable of integrating multiple wireless functions within a unified framework. For example, \cite{liu2025llm4wm} investigated multi-task fine-tuning for channel-related tasks, \cite{wen2025icwlm} introduced a wireless-native model leveraging in-context learning for multi-user precoding and channel prediction, and \cite{zheng2025muse} incorporated environmental awareness into a multi-task foundation model. 

While these advances demonstrate the promise of large models for wireless intelligence, they remain limited by two fundamental challenges -- 1) \textit{Modality constraint}: they often rely on aligning or converting wireless-domain data into the textual input space of pre-trained LLMs—an indirect cross-modal adaptation that leads to loss of representation fidelity and limited scalability across heterogeneous wireless tasks; 2) \textit{Task limitation}: current multi-task designs largely focus on channel-centric objectives, neglecting broader sensing dimensions such as human activity recognition, environmental perception, and multi-modal fusion. Addressing these challenges requires a new class of FMs capable of natively representing and generalizing across diverse wireless modalities and tasks.

To address these gaps, we propose \textbf{MMSense}, a multi-modal, multi-task wireless foundation model that unifies \emph{channel sensing}, \emph{human sensing}, and \emph{environment sensing} within a single cohesive framework. Our motivation arises from the observation that these seemingly distinct tasks are inherently correlated, as they all originate from and are governed by the same underlying radio-frequency (RF) signals that carry both communication and perceptual information. For instance, channel sensing tasks such as beam prediction not only characterize propagation dynamics but also implicitly encode information about user presence, mobility, and activity patterns. Likewise, human sensing leverages subtle variations in RF signal reflections and scattering to infer motion, gestures, and behavioral context, inherently carrying rich perceptual and environmental cues. By integrating multiple sensing modalities, including image, radar, and LiDAR data, our framework captures complementary representations that enhance both robustness and adaptability of predictions. Concurrently, the multi-task learning module facilitates the extraction of shared channel-centric knowledge across tasks, improving learning efficiency and generalization performance. 
Furthermore, modality embeddings produced by individual modality encoders are fused within the cross-attention layers of a backbone Vision-based LLM, where they dynamically interact with instruction tokens in the text branch to enable context-aware, multi-modal reasoning.

The contributions of this paper are summarized as follows:
\begin{itemize}
    \item We develop a unified framework that jointly addresses channel sensing, human sensing, and environment sensing through multi-modal data fusion, substantially extending the task scope beyond prior channel-centric approaches. Unlike existing methods that homogenize inputs by converting all modalities into a single representation, our framework \textit{organically} integrates heterogeneous data sources, including image, radar, LiDAR, and textual modalities, to preserve their complementary structures and maximize cross-domain information synergy.
    \item 
    We introduce a modality-adaptive gating scheme that dynamically weights multi-modal features according to their contextual reliability and task relevance, enabling the model to selectively emphasize the most informative modalities under varying sensing conditions. In parallel, a task-specific sequential attention module progressively refines signal feature representations across backbone FM layers, capturing hierarchical channel dependencies and task-aware feature evolution to achieve fine-grained adaptation for each sensing objective.
    \item Extensive experiments on the real-world datasets show that our method consistently outperforms existing baselines, demonstrating strong cross-modality generalization and robustness to diverse wireless sensing scenarios.
\end{itemize}

\section{Problem Formulation and Task Description}

In this section, we define the problem setting and outline the tasks addressed in this work. Specifically, we focus on three representative sensing categories that collectively characterize wireless perception: channel sensing, human sensing, and environment sensing.

\subsection{Channel-Centric Sensing}
For channel sensing, we focus on the beam prediction task, which is fundamental in high-frequency wireless systems such as mmWave and THz communication. In these systems, accurate beam alignment is essential to overcome severe path loss and maintain reliable connectivity, making beam prediction a key enabler for efficient channel utilization and robust link performance.
The objective of beam prediction is to infer the optimal beam index based on observed multi-modal data. 
Formally, given multi-modal inputs $\mathbf{X}$, the model outputs the beam index $\hat{b} \in \{1,2,\ldots,B\}$ that maximizes the achievable spectral efficiency. This is formulated as a multi-class classification problem:
\begin{equation}
    \hat{b} = \arg\max_{b \in \{1,\ldots,B\}} P(b \mid \mathbf{X}; \theta),
\end{equation}
where $P(b \mid \mathbf{X}; \theta)$ is the predicted probability distribution over beam indices parameterized by $\theta$. 

\subsection{Human-Centered Sensing}
Human sensing plays a central role in enabling human-centric 6G applications such as smart environments, healthcare, and human--machine interaction. 
In this sensing category, we consider two representative tasks: human activity recognition (HAR) and human pose estimation (HPE). 

HAR can be formulated as a multi-class classification problem. 
Given input $\mathbf{X}$, the model predicts the activity label $\hat{y}_h \in \{1,\ldots,C_h\}$:
\begin{equation}
    \hat{y}_h = \arg\max_{c \in \{1,\ldots,C_h\}} P(y_h = c \mid \mathbf{X}; \theta),
\end{equation}
where $C_h$ is the number of activity classes. As a regression problem, HPE aims to predict the 3D positions of keypoints for a human subject based on multi-modal RF signal inputs. 
Formally, given the same input $\mathbf{X}$, the model outputs the predicted keypoints:
\begin{equation}
    \hat{\mathbf{y}}_p = f_{\text{HPE}}(\mathbf{X}; \theta), \quad 
    \hat{\mathbf{y}}_p = [\hat{\mathbf{p}}_1, \hat{\mathbf{p}}_2, \ldots, \hat{\mathbf{p}}_J], 
    \quad \hat{\mathbf{p}}_j \in \mathbb{R}^d,
\end{equation}
where $f_{\text{HPE}}(\cdot)$ is the model function parameterized by $\theta$, $J$ is the number of keypoints, and $d$ is the coordinate dimension.

\subsection{Environment-Aware Sensing}
Moving blockages, such as pedestrians, vehicles, or other dynamic obstacles, pose a major challenge to high-frequency wireless links by inducing rapid fluctuations and sudden drops in signal quality. Unlike conventional static blockage detection, our task focuses on predicting future environmental changes caused by moving obstacles, inferred from variations in beam patterns and link quality between transceivers.
Formally, given current multi-modal observations $\mathbf{X}_t$ at time $t$, the objective is to predict whether the link will experience disruption at a future time $t + \Delta$. By forecasting \textit{future} blockage events rather than merely detecting existing ones, this task enables proactive situational awareness to support critical communication functions such as beam management, handover planning, and resource allocation in dynamic environments. We formulate this as a binary classification problem:
\begin{equation}
    \hat{y}_{e,t+\Delta} = \mathbb{I}\left[ P(y_{e,t+\Delta} = 1 \mid \mathbf{X}_t; \theta) > 0.5 \right],
\end{equation}
where $y_{e,t+\Delta} = 1$ indicates that a blockage will occur at time $t+\Delta$, and $\mathbb{I}[\cdot]$ is the indicator function that returns 1 if its argument is true and 0 otherwise.

\begin{figure*}[t]
	\centering
	\includegraphics[scale=0.75]{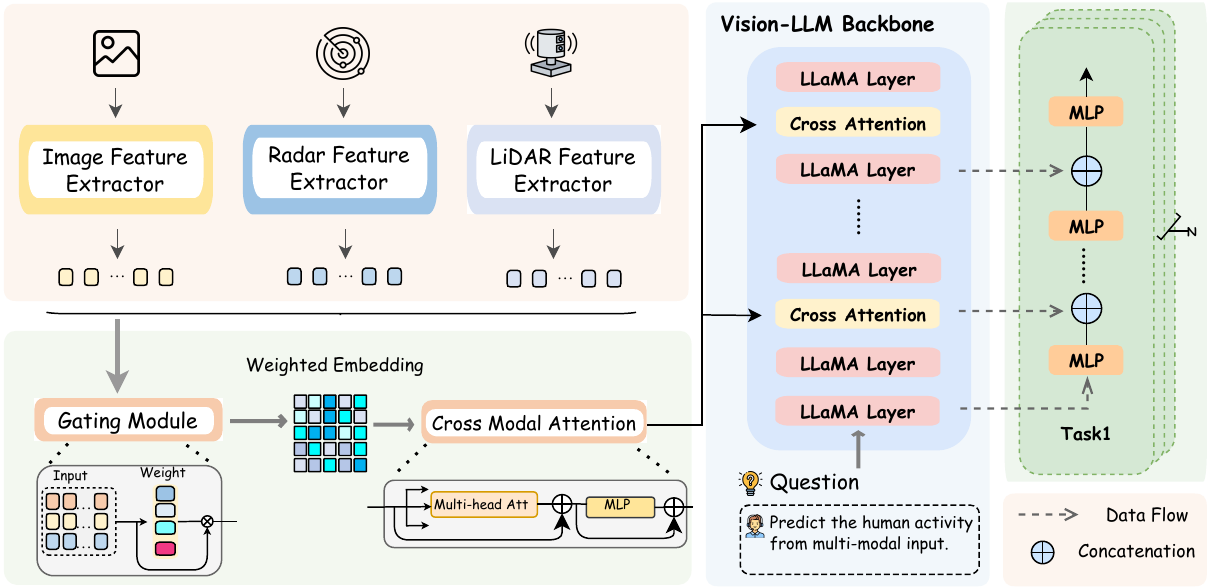}
	\caption{MMSense Framework with interrelated modules: (i) Modality-specific feature encoder; (ii) Modality-aware gating mechanism; (iii) Vision-LLM Backbone; (iv) Task-specific multi-Layer attention modules.}
	\label{framework}
    \vspace{-6mm}
\end{figure*}

\subsection{Unified Multi-Task Formulation}

While each of the above tasks targets a distinct sensing objective, they are inherently correlated through the shared radio environment and object-level contextual information. To exploit these cross-task dependencies, we formulate them within a unified multi-task learning framework. Formally, given a multi-modal input $\mathbf{X}$, the model jointly predicts the outputs for all sensing tasks as follows:
\begin{equation}
    \hat{\mathbf{y}} = f_{\text{MMSense}}(\mathbf{X}; \theta) = \{\hat{b}, \hat{y}_h, \hat{\mathbf{y}}_p, \hat{y}_e\},
\end{equation}
where $\hat{b}$ denotes the beam index, $\hat{y}_h$ is the human activity label, $\hat{\mathbf{y}}_p$ is the predicted keypoint coordinates, and $\hat{y}_e$ represents future blockage prediction.  
To optimize the model parameters $\theta$, we formulate a joint learning objective that enables shared representation learning across related sensing tasks while preserving task-specific specialization. The overall optimization objective is defined as a weighted sum of task-specific losses:
\begin{equation}
    \min \{\mathcal{L}_{\text{total}} = \sum_{t \in \{\text{BP, HAR, HPE, FBP}\}} \lambda_t \, \mathcal{L}_t \},
\end{equation}
where $\lambda_t$ are task-specific weights that balance the contribution of each task in a shared radio environment. 

\section{MMSense: Multi-task Multi-modal Sensing}

To address the optimization problem defined in Eq.~(6), we introduce the MMSense framework as a unified solution. The overall architecture are illustrated in Fig.~\ref{framework}.

\subsection{Modality-Specific Feature Encoding}
Each sensing modality exhibits unique structural characteristics, necessitating specialized feature extractors. 
Let $X_m$ denote the input of modality $m \in \{\text{image}, \text{radar}, \text{LiDAR}\}$. 
We develop pre-trained modality-specific encoders $F_m(\cdot; \theta_{F,m})$ to extract high-dimensional representations defined as:
\begin{equation}
    z_m = F_m(X_m; \theta_{F,m}).
\end{equation}
\noindent For the image modality, which contains rich spatial and semantic cues, we adopt a ResNet backbone to capture hierarchical visual features that effectively represent scene-level context and object appearance.
For radar and LiDAR modalities, which provide geometric and structural information of the physical wireless environment, we employ Point Transformer encoders~\cite{zhao2021point} to model fine-grained 3D spatial relations among point clouds. This design preserves geometric fidelity and enables robust feature learning under varying signal conditions. In this way, the framework ensures that each sensing input is transformed into a semantically aligned, high-dimensional representation space suitable for subsequent joint learning.

\subsection{Modality-Aware Gating Mechanism}
While each modality contributes complementary information, their relative reliability varies across sensing conditions.
For instance, radar tends to perform better in low-light or adverse weather scenarios, whereas LiDAR remains more effective under visual occlusion. Consequently, assigning \textit{equal} importance to all modalities can lead to suboptimal fusion and degraded overall performance. To overcome this limitation, we introduce a modality-adaptive gating network, inspired by the \textit{mixture-of-experts} paradigm, which dynamically evaluates the contextual relevance of each modality and generates adaptive fusion weights to achieve more effective and robust multi-modal integration.
The gating network $g(\cdot; \theta_g)$ computes modality weights:
\begin{equation}
    [w_{\text{img}}, w_{\text{radar}}, w_{\text{LiDAR}}] 
    = \text{softmax}(g([z_{\text{img}}, z_{\text{radar}}, z_{\text{LiDAR}}]; \theta_g)),
\end{equation}
where $\sum_m w_m = 1$ and $w_m \geq 0$, and $z_{(\cdot)}$ is derived from Eq. (7).  
This gating network first aggregates multi-modal extracted features into a compact intermediate representation that captures cross-modality interactions and dependencies. Such representation is then mapped into modality-specific scores through trainable neural layers. Next, the scores are converted into normalized weights via the softmax operation, explicitly reflecting the instantaneous importance of each sensing modality.

\subsection{Cross-Modal Attention Fusion}

After obtaining the modality-specific embeddings and weighting them via the gating mechanism, the next step is to integrate them into a unified cross-modal representation suitable for various downstream tasks. 
Simply concatenating or averaging these weighted embeddings is insufficient, as it fails to capture complex dependencies across modalities. To address this, we design a cross-modal attention module, which explicitly models information exchange between modalities while preserving distinctive features.
Let $\{f_m\}_{m=1}^D$ denote the gated embeddings from $D$ modalities, where $f_m=w_mz_m$. 
We concatenate them to form a multi-modal sequence as:
\begin{equation}
    b_{\text{mm}} = [f_{\text{img}}; f_{\text{radar}}; f_{\text{LiDAR}}] \in \mathbb{R}^{(n_f \cdot D) \times d},
\end{equation}
where $n_f$ is the number of tokens per modality and $d$ is the embedding dimension. 
The sequence is then linearly projected into queries, keys, and values:
\begin{equation}
    \{Q_{\text{mm}}, K_{\text{mm}}, V_{\text{mm}}\} = \{b_{\text{mm}} W_Q, b_{\text{mm}} W_K, b_{\text{mm}} W_V\}.
\end{equation}
Next, the multi-head dot-product attention is applied to model dependencies among modality features as follow:
\begin{equation}
    \text{Attn}(Q_{\text{mm}}, K_{\text{mm}}, V_{\text{mm}}) = \text{softmax}\left(\frac{Q_{\text{mm}} K_{\text{mm}}^\top}{\sqrt{d}}\right)V_{\text{mm}}.
\end{equation}
By combining the gating mechanism with the cross-modal attention operation, the model learns to emphasize the most \textit{reliable} modality while still aligning complementary features. 

\subsection{Vision-LLM Backbone Integration}
After obtaining the fused cross-modal representation $b_{\text{mm}}$, we feed it into a vision–language model backbone (such as the LLaMA model architecture from~\cite{touvron2023llama}). This design choice is motivated by two key considerations. First, Vision-LLMs are pre-trained on large-scale multi-modal datasets, endowing them with strong cross-modal alignment and reasoning capabilities. Second, the cross-attention layers within the Vision-LLM naturally align our fused multi-modal embeddings with the textual embedding space, enabling natural language supervision and instruction tuning. This integration introduces an additional modality—language—allowing the model to benefit from human-interpretable guidance and \textit{continual adaptation} across diverse sensing and reasoning tasks.

For instance, we introduce task-specific prompts that describe the sensing objective -- “\textit{Predict the human activity from multi-modal inputs}”. These textual instructions are tokenized and combined with the previously fused embedding $b_{\text{mm}}$ through the Vision-LLM's cross-attention layers, enabling interaction between semantic language tokens and multi-sensor representations.
Moreover, we apply low-rank adaptation (LoRA) fine-tuning to efficiently adapt the Vision-LLM to multi-modal sensing data without retraining the full backbone model. This lightweight adaptation injects low-rank updates into attention layers, preserving language knowledge while specializing the model for sensing-related tasks.

As a result, the Vision-LLM backbone unifies heterogeneous sensing modalities in a text-aligned space, and enables multi-task learning under a shared instruction-following paradigm. 

\subsection{Task-Specific Multi-Layer Attention Modules}

As the last step, to effectively exploit the hierarchical representations within the Vision-LLM backbone, we do not directly take the \textit{final hidden state} for all sensing tasks. 
Instead, we design task-specific attention modules that progressively aggregate information from multiple \textit{intermediate} layers. 

Let $\{H^l\}_{l=1}^L$ denote the hidden states from the $L$ transformer layers of the Vision-LLM backbone. 
For each downstream task $t$, we attach a sequence of attention modules $\{\mathcal{A}^t_1, \mathcal{A}^t_2, \ldots, \mathcal{A}^t_K\}$, each operating on one hidden state. 
The output of one module is concatenated with the next layer’s hidden state and passed to the next module as follows:
\begin{align}
    z^t_1 &= \mathcal{A}^t_1(H^{l_1}), \\
    z^t_k &= \mathcal{A}^t_k([z^t_{k-1}; H^{l_k}]), \quad k = 2, \ldots, K,
\end{align}
where $[\,\cdot\,;\,\cdot\,]$ denotes concatenation along the feature dimension. Specifically, each attention module $\mathcal{A}^t_k(\cdot)$ is implemented as a lightweight two-layer MLP that learns an adaptive weighting over its input features. This module effectively reweights the importance of sensing features and captures \textit{task-relevant cues} before passing them to the next stage.

To jointly optimize all tasks while balancing their relative importance, we adopt an uncertainty-based multi-task loss weighting strategy. Instead of manually tuning task-specific coefficients, the model learns adaptive loss weights based on the homoscedastic uncertainty associated with each task~\cite{kendall2018multi}. 

Let $\mathcal{L}_t$ denote the loss function for task $t $.
Each task is associated with a learnable log-variance parameter $\sigma_t^2$, representing the observation uncertainty. Thus,  
the cross-task training objective becomes:
\begin{equation}
    \mathcal{L}_{\text{total}} 
    = \sum_{t} \frac{1}{2\sigma_t^2} \mathcal{L}_t + \log \sigma_t.
\end{equation}
Here, $\frac{1}{2\sigma_t^2}$ adaptively scales each task’s contribution according to its uncertainty, while the $\log \sigma_t$ term regularizes the optimization to prevent trivial solutions.

\section{Experiments and Validation}

In this section, we evaluate the performance of the proposed MMSense. We first introduce the datasets used for our experiments, followed by a description of the baseline methods. We then present quantitative results to validate the effectiveness of our model across multiple sensing tasks.

\subsection{Datasets and Platform}

First, the MM-Fi dataset~\cite{yang2023mm} provides a large-scale benchmark for human sensing tasks. It includes synchronized multi-modal data collected from RGB cameras, radar sensors, and LiDAR scanners, covering a wide range of human activities and poses. We utilize the dataset for our two human sensing tasks. The first task is the human activity recognition task, where a total number of 27 activity categories is included. For the human pose estimation task, the dataset contains over 320,000 synchronized frames collected from 40 human subjects. The ground-truth labels include 17 key human joint positions.
Second, the DeepSense 6G dataset \cite{DeepSense} is an extensive multi-modal dataset designed for wireless research.  It provides co-located measurements of communication and sensing modalities, including mmWave channel state information, radar range–angle maps, LiDAR scans, and synchronized camera images across various indoor and outdoor environments. We employ this dataset for our channel and environment sensing tasks. The MMSense model is trained with 1 A100 GPU using NVIDIA Saturn Cloud platform.

\subsection{Baseline Schemes and Evaluation Metrics}
To validate the effectiveness of the proposed framework, we compare our model with several learning-based methods. We first compare our model with the single-task small models. These models are dedicated to a specific sensing task, and they have relatively small model parameters.
\begin{itemize}
    \item \textbf{MLP-based model}\cite{alrabeiah2020deep}:  A multi-layer perceptron trained on the flattened input features. This model serves as a simple baseline to assess the non-linear fitting capability.
    \item \textbf{CNN-based model}\cite{safari2019deep}:  CNN-based methods mainly utilize local convolutional kernels to extract spatially localized features from image-like modalities.
    \item \textbf{Transformer-based model}\cite{vaswani2017attention}: This scheme leverages self-attention mechanisms to model global dependencies among features. Unlike CNNs, they do not rely on local receptive fields, allowing them to capture long-range spatial or temporal relationships across modalities.
    
\end{itemize}

Besides the small models that focus on a single task, we also compare our method with state-of-the-art large models: 
\begin{itemize}
    \item \textbf{LLM4WM}\cite{liu2025llm4wm}:  This is the first work that applies LLM as backbone to solve multiple tasks in wireless networks. We adapt this framework to the multi-modal setting for a fair comparison with our method.
\end{itemize}

For each sensing task, we employ task-specific evaluation metrics. Specifically, for human activity recognition, beam prediction, and blockage prediction, which are formulated as classification tasks, we use Top-1 accuracy to assess performance. For human pose estimation, which involves continuous keypoint regression, we use the Mean Squared Error (MSE) to evaluate the precision of the predicted joint coordinates.

\begin{table}[!t]
\centering
\caption{Performance comparison of our method and baselines across four sensing tasks. 
The best result is \textbf{boldfaced}, and the second best is \underline{underlined}.}
\label{tab:main_results}
\renewcommand{\arraystretch}{1.2}
\setlength{\tabcolsep}{3.8pt} 
\begin{tabular}{lcccc}
\toprule
\textbf{Method} &
\makecell{\textbf{Beam}\\\textbf{Prediction} $\uparrow$} &
\makecell{\textbf{HAR} $\uparrow$} &
\makecell{\textbf{HPE} $\downarrow$} &
\makecell{\textbf{Blockage}\\\textbf{Prediction} $\uparrow$} \\
\midrule
MLP          & 0.0656 & 0.8491 & 0.0268 & 0.9403 \\
CNN          & 0.1438 & 0.8583 & 0.0177 & 0.9326 \\
Transformer  & \underline{0.1972} & 0.8550 & 0.0151 & 0.9378 \\
LLM4WM       & 0.1504 & \textbf{0.8816} & \underline{0.0119} & \underline{0.9433} \\
\textbf{Ours} & \textbf{0.3942} & \underline{0.8766} & \textbf{0.0058} & \textbf{0.9453} \\
\bottomrule
\end{tabular}
\end{table}

\subsection{Accuracy of Sensing Tasks}

Table~\ref{tab:main_results} reports the performance comparison of our method and four baselines across the all sensing tasks. 
It is observed that our approach achieves the best results in three tasks and competitive performance in the remaining one, demonstrating its strong sensing capability across diverse tasks.
For the beam prediction task, our method attains the highest accuracy, substantially surpassing all baselines. 
This validates the effectiveness of our modality-aware gating mechanism and cross-modal attention fusion in exploiting complementary radar and LiDAR information.
In human sensing tasks, our model achieves the best performance in HPE, while the HAR accuracy is slightly below the top-performing baseline. This can be attributed to HAR being a high-level semantic task where image features alone capture most discriminative cues, thus limiting the marginal gains from the multi-modal fusion.
For the blockage prediction task, our method again achieves the highest accuracy, reflecting improved environmental awareness and robustness enabled by the multi-task learning design.


In detail, Table~\ref{tab:ablation_results} presents the ablation study of our MMsense to analyze the impact of two design components: the modality-aware gating mechanism and the task-specific multi-layer attention.
First, removing the gating mechanism leads to noticeable performance drops in both beam prediction and blockage prediction tasks, confirming that adaptive modality weighting plays a critical role in managing modality reliability variations under diverse sensing conditions. Second, removing the task-specific multi-layer attention leads to a significant degradation in HPE performance, confirming that our sequential attention design is essential for refining hierarchical feature representations. Interestingly, HAR accuracy improves in this ablation setting, as high-level activity recognition primarily depends on image semantics. Aggregating multiple hidden states from the LLM may introduce redundant information or noise, whereas removing this component enables the model to focus on the most discriminative features, thereby preserving semantic clarity and improving classification accuracy.
Overall, the full MMSense model achieves the consistently high performance across all tasks, demonstrating that both components contribute synergistically to improving the multi-modal and multi-task sensing capability.

\begin{table}[!t]
\centering
\caption{Ablation study of our proposed framework.}
\label{tab:ablation_results}
\renewcommand{\arraystretch}{1.15}
\setlength{\tabcolsep}{3pt} 
\small 
\begin{tabular}{lcccc}
\toprule
\textbf{Method} &
\makecell{\textbf{Beam}\\\textbf{Prediction} $\uparrow$} &
\makecell{\textbf{HAR} $\uparrow$} &
\makecell{\textbf{HPE} $\downarrow$} &
\makecell{\textbf{Blockage}\\\textbf{Prediction} $\uparrow$} \\
\midrule
w/o Gating               & 0.3272 & 0.8386 & \underline{0.0068} & 0.9378 \\
w/o Task-Att         & \underline{0.3785} & \textbf{0.8807} & 0.1152 & \underline{0.9410} \\
\textbf{Full Model}           & \textbf{0.3942} & \underline{0.8766} & \textbf{0.0058} & \textbf{0.9453} \\
\bottomrule
\end{tabular}
\end{table}

\subsection{Generalization Ability Across Sensing Tasks}
\begin{figure}[htbp]
\centerline{\includegraphics[scale=0.32]{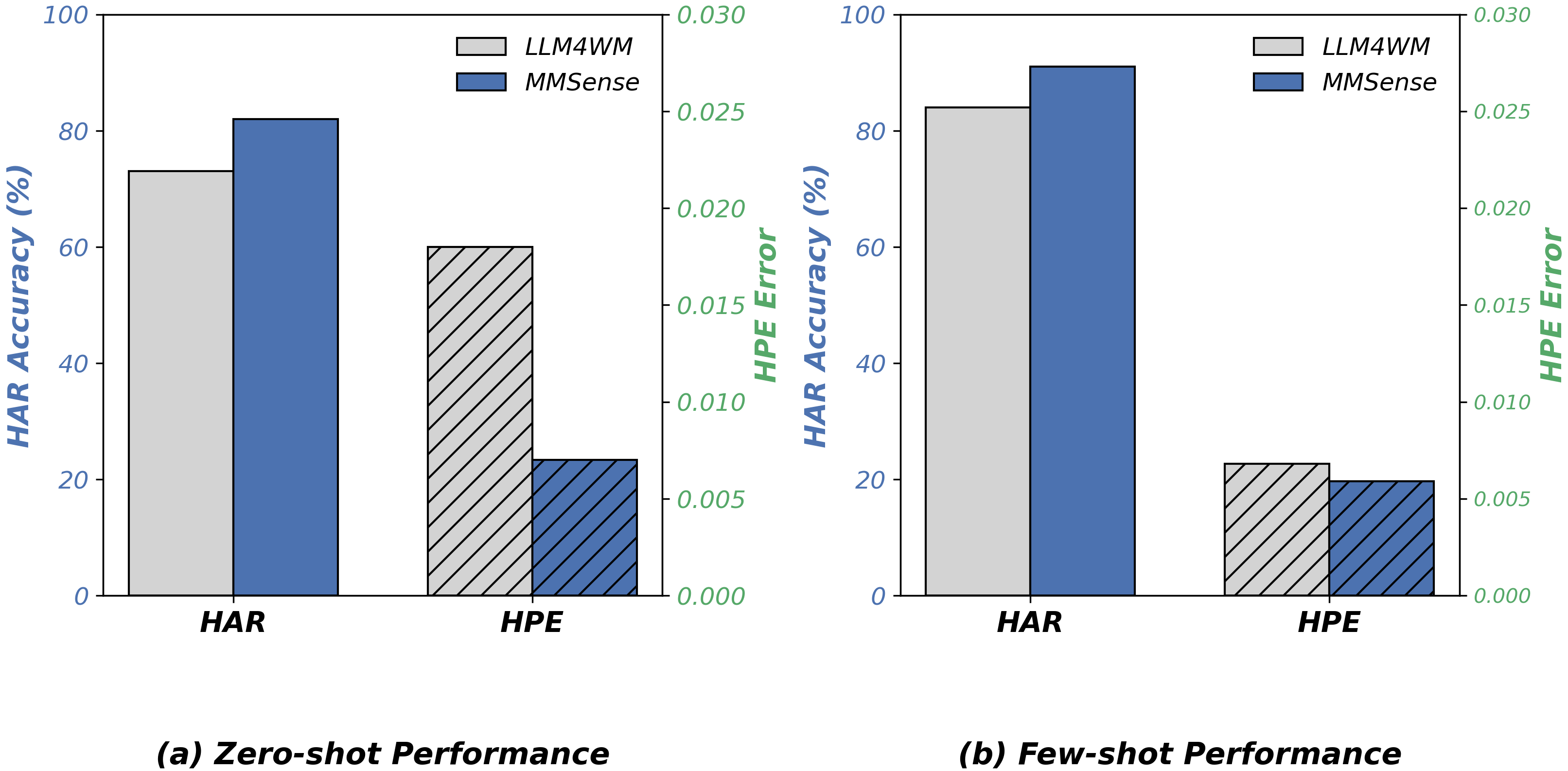}}
\caption{Generalization performance of our model and the baseline. }
\label{gen}
\vspace{-3mm}
\end{figure}

To evaluate the generalization capability of the proposed MMSense framework, we investigate its zero-shot and few-shot performance on the human sensing tasks, as shown in Fig.~\ref{gen}. In the zero-shot setting, the model is tested on actions performed by unseen human subjects. MMSense achieves noticeably higher HAR accuracy and significantly lower HPE error compared to the baseline model, demonstrating its strong ability to transfer knowledge across heterogeneous modalities and tasks without additional fine-tuning. This also highlights the effectiveness of the proposed cross-modal fusion and task-specific attention mechanisms in capturing shared semantic structures and enhancing generalization to unseen scenarios.

In the few-shot setting, where only 10\% of labeled samples are available for fine-tuning, MMSense further improves both HAR and HPE performance, surpassing the baseline across all evaluation metrics. The performance gain is particularly notable in HPE, indicating that MMSense effectively leverages limited supervision to refine fine-grained spatial and structural understanding of human motion. 
These zero- and few-shot results further reveal that MMSense effectively leverages the Vision-LLM backbone’s reasoning and generalization capabilities in complex wireless sensing environments.

\section{Conclusion}

In this paper, we propose MMSense, a multi-modal, multi-task foundation model that unifies channel, human, and environment sensing within a cohesive framework. By integrating image, radar, and LiDAR data into a vision-aligned embedding space, our approach enables effective cross-modal feature alignment and instruction-driven task adaptation. The proposed modality-aware gating mechanism and task-specific multi-layer attention modules allow the model to dynamically prioritize relevant modalities and refine hierarchical representations for each task. Extensive experiments on real-world datasets demonstrate that MMSense consistently outperforms both single-task and existing large-model baselines in sensing accuracy and task generalization performance. 

\bibliographystyle{IEEEtran}
\bibliography{reference}

\end{document}